\documentclass[journal]{IEEEtran}
\pdfoutput=1

\usepackage{cite}
\usepackage{color}

\usepackage{graphicx}
\usepackage{multirow, multicol}
\usepackage[table,xcdraw]{xcolor}

\usepackage{amsmath}
\usepackage{amsthm}
\usepackage{amssymb,mathtools}

\usepackage{cuted}
\usepackage{siunitx}
\usepackage{array}

\usepackage{color, colortbl}
\usepackage{xcolor}
\definecolor{lightgray}{gray}{0.1}
\definecolor{orange}{rgb}{1,0.5,0}
\definecolor{blue}{rgb}{0,0.,1}

\usepackage[caption=false,font=normalsize,labelfont=sf,textfont=sf]{subfig}

\begin{document}

\title{Shareable Driving Style Learning and Analysis with a Hierarchical Latent Model}

\author{Chaopeng~Zhang$^1$,      
        Wenshuo~Wang$^2$,~\IEEEmembership{Member,~IEEE,}
        Zhaokun~Chen$^1$,
        Jian~Zhang$^3$,
        Lijun Sun$^4$,~\IEEEmembership{Senior Member,~IEEE,}
        and~Junqiang~Xi$^1$
\thanks{This paper was received by xxx, revised by xxxx. This work was supported by the National Natural Science Foundation of China (Grant No.: 52272411) and the Canada IVADO Postdoctoral Fellowship. (Corresponding Authors: Wenshuo Wang and Junqiang Xi)}
\thanks{$^1$C. Zhang, J. Xi, and Z. Chen are with the School of Mechanical Engineering, Beijing Institute of Technology, Beijing, China.
         {\tt\small chaopeng666@gmail.com; xijunqiang@bit.edu.cn; zhaokunchen184@gmail.com}}%
\thanks{$^2$W. Wang is with the School of Mechanical Engineering, Beijing Institute of Technology, Beijing, China. He was also with the Department of Civil Engineering, McGill University, Montreal, QC, Canada.
         {\tt\small wwsbit@gmail.com}}%
\thanks{$^3$J. Zhang is with the China FAW Group Co., Ltd., Changchun, China.
         {\tt\small zhangjian3@faw.com.cn}}%
\thanks{$^4$L. Sun is with the Department of Civil Engineering, McGill University, Montreal, QC, Canada.
         {\tt\small lijun.sun@mcgill.ca}}%
}

\markboth{Journal of \LaTeX\ Class Files,~Vol.~14, No.~8, August~2015}%
{Shell \MakeLowercase{\textit{et al.}}: Bare Demo of IEEEtran.cls for IEEE Journals}

\maketitle

\begin{abstract}
Driving style is usually used to characterize driving behavior for a driver \textit{or} a group of drivers. However, it remains unclear how one individual's driving style shares certain common grounds with other drivers. Our insight is that driving behavior is a sequence of responses to the weighted mixture of latent driving styles that are shareable \textit{within} and \textit{between} individuals. To this end, this paper develops a hierarchical latent model to learn the relationship between driving behavior and driving styles.  We first propose a fragment-based approach to represent complex sequential driving behavior, allowing for sufficiently representing driving behavior in a low-dimension feature space. Then, we provide an analytical formulation for the interaction of driving behavior and shareable driving style with a hierarchical latent model by introducing the mechanism of Dirichlet allocation. Our developed model is finally validated and verified with 100 drivers in naturalistic driving settings with urban and highways. Experimental results reveal that individuals share driving styles within and between them. We also analyzed the influence of personalities (e.g., age, gender, and driving experience) on driving styles and found that a naturally aggressive driver would not always keep driving aggressively (i.e., could behave calmly sometimes) but with a higher proportion of aggressiveness than other types of drivers. 
\end{abstract}


\IEEEpeerreviewmaketitle


\section{Introduction}
\IEEEPARstart{H}{igh} acceptance of driver assistance systems requires adapting to individuals' driving style \cite{belcher2022eeg,wang2019improving}, such as steering assistance systems \cite{wang2016human}, active suspension control \cite{wang2015study}, adaptive cruise control \cite{wang2017development,gao2020personalized,zhu2019typical}, lane departure warning systems \cite{wang2018learning}, and eco-driving mode \cite{yang2018driving}. Many researchers directly labeled or recognized individual drivers with a \textit{fixed} personality, such as aggressive or calm. However, driving styles can be influenced by personalities (e.g., age, education experience, gender, and social preference) and traffic environments (e.g., traffic flow, surrounding agents, and weather), and could be changed immediately over space and time. For instance, a naturally calm driver could \textit{temporarily} become aggressive when tending to reach their specific driving goals emotionally, such as driving to the emergency center. Therefore, a human's driving style in the real world should be dynamically changeable and evolve across their driving behaviors and traffic environment. 

There exist different arguments for the changeable attributes of driving style. Some researchers believe that driving style is a habitual way of driving \cite{sagberg2015review} and is a stable character (e.g., driving attitude \cite{ishibashi2007indices,elander1993behavioral}) over a long-term period. Most existing works treat an individual's driving style as fixed attributes. On the contrary, some researchers treat driving style as a response of \textit{instantaneous} preference in a personal way, i.e., a function of many factors, such as environmental awareness and driving intent, skills, and willingness to take risks. For example, a single driver's driving style is inconsistent and may vary even within a single trip \cite{feng2018driving}. Moreover, not only does driving style vary between individuals, but the thresholds for aggressive and calm driving vary across roadway types \cite{mohammadnazar2021classifying} and driving tasks (i.e., cruising and ride request) \cite{ma2021driving}. 

Driving styles are application-dependent and can be categorized into finitely countable groups or captured using continuous metrics. For example, the continuous metrics for driving style are mainly utilized for eco-driving systems \cite{manzoni2010driving, corti2013quantitative, neubauer2013accounting}, while some researchers categorized driving styles into discrete classes (e.g., calm, normal, and aggressive) using supervised \cite{karginova2012data,xu2015establishing,vaitkus2014driving,augustynowicz2009preliminary}, unsupervised \cite{constantinescu2010driving,miyajima2007driver,ishibashi2007indices,wang2011review}, and semi-supervised \cite{wang2017driving} classification or clustering techniques with the assumption that individual driving styles remain unchanged during driving.  The discrete classification provides a semantic way to understand and analyze complex driving behaviors. Wang \cite{wang2018driving} discretized driving behavior into primitive patterns to analyze driving style semantically, which allows the authors to evaluate the similarity between driving behaviors using a probabilistic measure \cite{wang2019probabilistic}. The discretization of driving styles also informs the transition probabilities of driving maneuver states \cite{li2017estimation} and is suitable for some hierarchical latent models \cite{qi2015leveraging, qi2015appropriate}. The aforementioned works highlight the dynamic process and distribution of individuals' driving behavior patterns (i.e., the semantic abstraction of driving behaviors) and their distribution in specific scenarios, such as car-following or lane-changing, but fail to analyze the shareable characteristics of driving styles within and between individuals.

On the one hand, human drivers should share some common ground regarding driving style because hundreds of millions generate diverse driving behaviors only with finite categories of explainable driving styles (e.g., aggressive, normal, and calm). For example, a naturally aggressive driver distributively shares some common grounds of driving styles with a naturally calm driver. On the other hand, an individual driver could behave in different driving styles at different times and in traffic scenarios.  It remains open to how these styles are distributed \textit{within} and \textit{between} individuals.  Providing insights into the underlying relationship between driving style and driving behavior in a computational way can uncover the shareable attributes among individual drivers. Unlike existing works that independently identify a fixed driving style for individuals, we aim to learn a set of shareable driving styles for a group of drivers based on their driving operation sequences. To this end, we treat driving styles as shareable latent states, use the fragments of driving operation sequences over time as the responses of driving styles, and develop a hierarchical latent model to learn these driving styles. 

In summary, unlike existing work treating driving styles as a deterministic feature for individuals, we propose a novel perspective that \textit{an individual's driving behavior is a mixture of weighted driving styles in a probabilistic way}. The main contributions of this paper are twofold:
\begin{enumerate} 
    \item Propose an analytically fragment-based hierarchical latent model that can capture the within- and between-shareable attributes of driving styles.
    \item Learn and analyze the differences in the shareable driving styles with respect to the personalities of 100 drivers in naturalistic driving settings.
\end{enumerate}

The remains of this paper are organized as follows. Section II introduces our proposed fragment-based hierarchical latent model. Section III describes driving data collection and experimental setting. Section IV  presents the experiment results and analysis, followed by the conclusions and further work in Section V.

\section{Fragment-based Hierarchical Latent Model}
Our proposed model consists of four parts: driving behavior representation, hierarchical latent model, model learning algorithms, and evaluation. In what follows, we will detail each part.
\subsection{Driving Behavior Representation}
\label{subsec:representation}

\begin{figure}[t]
    \centering
    \includegraphics[width=\linewidth]{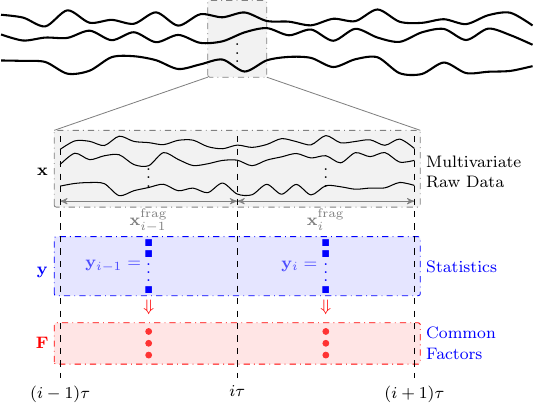} 
    \caption{The procedure of driving behavior representation from a continuous high-dimension observation sequence $\{\mathbf{x}\}$ to a discrete low-dimension representative common factors $\mathbf{F}$.}
    \label{fig: driving behavior representation}
\end{figure}

The driving behavior representation procedure consists of behavior fragmentation and dimension reduction, as illustrated in Fig. \ref{fig: driving behavior representation}.
\subsubsection{Behavior fragmentation}
The driving style implicit in the driving behavior is usually rather stable for a short time (e.g., in seconds). The fragment could provide more informative driving style attributes than a single data sample. Therefore, we propose to utilize the fragment of sequential driving behavior as observations instead of using each sampling data point of sensory information.  
 
Given a sequence of continuous vector-valued sensory information $\{\mathbf{x}_{t}\}$ of driving behavior with sampling time step $t$ and $\mathbf{x}_{t}\in \mathbb{R}^{p\times 1}$, we segment it into finite fragments of equal length $\tau$ sequentially (see Fig. \ref{fig: driving behavior representation}). The $i$-th fragment is 

\begin{equation}
\mathbf{x}_{i}^{\mathrm{frag}} = \{ \mathbf{x}_{t}\}, \ \mathrm{for} \ t \in [(i-1)\tau,i\tau])
\end{equation}
where $\mathbf{x}_i^{\mathrm{frag}} \in \mathbb{R}^{p \times \tau}$ is a matrix-valued variable and $i = 1 ,2 , \dots, N$. Then a vector of statistics, $\mathbf{y}_{i}$, of each fragment is used to capture the driving behavior in the $i$-th fragment

\begin{equation}
\mathbf{y}_{i} = [\max\{ \mathbf{x}_{t} \},\mathrm{avg} \{ \mathbf{x}_{t} \} , \min \{ \mathbf{x}_{t} \} ]^{\top}
\end{equation}
for $t \in [(i-1)\tau,i\tau])$ with $\mathbf{y}_i \in \mathbb{R}^{3p\times 1}$ a vector-value variable. Each fragment's statistics --- maximum, minimum, and mean values can reflect the driving style quickly. 
 
\subsubsection{Dimension reduction}
\label{subsubsec:reduction}
The statistics $\mathbf{y}$ for each fragment are usually in a high dimension, preventing semantic analysis and interpretation. To solve this problem, we introduce factor analysis to transfer the high-dimension driving data into a low-dimensional space for efficient analysis by discovering some common factors influencing driving behavior statistics \cite{decoster1998overview} so that observed driving behavior statistics can be easily interpreted and understood \cite{yong2013beginner}. A fewer number of common factors would provide a much easier analysis \cite{mcdonald2014factor}. Therefore, we seek the common factors with the smallest number while accounting for the correlations between driving behavior statistics.

Given the statistics of each fragment of all driving behavior $\{\mathbf{y}_{i}\}_{i=1}^{N}$, we stack them sequentially and obtain a matrix  $\mathbf{Y}$ of which each row represents the statistic feature through all driving behavior fragments as 

\begin{equation}
\label{eq:reform}
\mathbf{Y}= 
\begin{bmatrix} 
| & | & \cdots & | \\
\mathbf{y}_1 & \mathbf{y}_2 & \cdots & \mathbf{y}_N \\
| & | & \cdots & | \\
\end{bmatrix}
=
\begin{bmatrix} 
\mathbf{y}^{(1)} \\
\mathbf{y}^{(2)}\\
 \vdots  \\
\mathbf{y}^{(3p)} \\
\end{bmatrix},
\end{equation}
where the $j$-th statistic feature is demoted as $\mathbf{y}^{(j)} \in \mathbb{R}^{N}$ for all features $j = 1, 2, \dots, 3p$. To reduce the dimension of the statistic features, we assume there exists $m$ (usually $m<3p$) underlying common factors $ \mathbf{f}^{(\ell)} \in \mathbb{R}^{N}$ whereby each statistic feature, $\mathbf{y}^{(j)}$, is a linear combination of the factors $\{\mathbf{f}^{(\ell)}\}_{\ell=1}^{m}$ with a residual variate $\mathbf{e}$, which can be rewritten into a matrix form
\begin{equation}
    \mathbf{Y} = \mathbf{A}\mathbf{F} + \mathbf{e}
\end{equation}
with
\begin{equation}
\mathbf{A} = 
\begin{bmatrix}
a_{11} & a_{12} & \cdots & a_{1m} \\
a_{21} & a_{22} & \cdots & a_{2m} \\
\vdots & \vdots & \ddots & \vdots \\
a_{3p,1} & a_{3p,2} & \cdots & a_{3p,m}
\end{bmatrix}
, \ 
\mathbf{F} = 
\begin{bmatrix}
\mathbf{f}^{(1)} \\
\mathbf{f}^{(2)} \\
\vdots \\
\mathbf{f}^{(m)}
\end{bmatrix},
\end{equation}
where $\mathbf{A}$ is the matrix of factor weights and each element $a_{j\ell}$ represents the contribution of $j$-th feature $\mathbf{y}^{(j)}$ to the $\ell$-th factor $\mathbf{f}^{(\ell)} \in \mathbb{R}^{N}$.  A larger value of factor weights indicates more contribution to that factor from the corresponding feature \cite{harman1976modern}. Here we notice that the $i$-th element of the factor $\mathbf{f}_{i} = [f^{(1)}_{i}, \dots, f^{(\ell)}_{i}, \dots, f^{(m)}_{i}]^{\top}$ corresponds to the $i$-th fragment, and thus each fragment can be represented using only $m$ factors. In such a way, the low-dimensional factors enable a semantic analysis of driving behavior for a specific driving style.

\subsection{Hierarchical Latent Driving Behavior Model}
\label{subsection: Hierarchical Latent Driving Behavior Model}
Based on the obtained fragments of driving behavior from the above section, we develop a hierarchical latent model to formulate human driving behavior concerning driving styles, which can be learned from sensory data directly. In what follows, we will detail the modeling procedures of driving styles, driving behavior, and their connections. 

\subsubsection{Driving style} Driving style depicts a habitual way of driving, characteristic of a driver or a group of drivers, which retains two main attributes:

\paragraph{\textbf{Unobservable}}
Driving style cannot be measured directly using physical sensors but can be inferred from the behavior observations (e.g., speed and acceleration). In addition, an individual's driving style could change over space and time with the influences of other factors such as traffic conditions. We treat driving style as a latent random variable, denoted as $z$. For practical purposes, driving style is usually interpreted in a discrete format, such as aggressive and calm, and thereby we  formulate it as a discrete random variable and assign each style with an integer index $k\in [1, K]$, where $K$ is the number of categories of driving styles.

\paragraph{\textbf{Shareable}} On the one hand, driving style is temporally self-shareable \textit{within} drivers. The driver could exhibit the same style at different moments and places of driving. We formulate the shareable driving style of an individual driver $d$ for single behavior fragments in the temporal space using a categorical distribution

\begin{equation}
\label{eq:zdCat}
z_d \sim \mathrm{Cat}(K,\boldsymbol{\theta}_d)
\end{equation}
with $\boldsymbol{\theta}_d = \left [\theta_{d1}, \dots, \theta_{dK} \right ]$ and $\sum_{k=1}^K \theta_{dk} = 1$, where $\theta_{dk}$ represents the probability of driving style being $k$ for driver $d$ (i.e., $z_{d} = k$). Assigning the categorical distribution to the driver's temporally-distinct driving behavior fragments realize sharing one identical driving style, which is called temporally self-shareable. Thus, we can use a sequence of random variables $\left \{ z_{d1}, \dots, z_{dN} \right \}$ to represent the sequence of driving styles for all driving behavior fragments. Note that $z_{di}$ is a random variable and can be one value of integers $ [1, K]$ representing driving styles. The count of each driving style for the sequence of observed driving behavior is formulated as the multinomial distribution

\begin{equation}
\label{eq:cdMulti}
\mathbf{c}_d \sim \mathrm{Multi}(N, \boldsymbol{\theta}_d)
\end{equation}
with $\mathbf{c}_d = \left[ c_{d1}, \dots, c_{dk}, \dots, c_{dK}\right]$, where $c_{dk}$ represents the count of the driving behavior fragments tagged with driving style $k$, and $N$ is the number of all fragments.

On the other hand, one specific label of driving style (e.g., aggressive) can characterize a group of drivers \cite{sagberg2015review}. In other words, the driving style is shareable \textit{between} individual drivers. To formulate this attribute, we treat the parameter $\boldsymbol{\theta}_{d}$ as a vector-valued random variable subject to a specific distribution to indicate the shared information regarding driving style (i.e., proportions of each driving style) between individuals. Here, we select a Dirichlet distribution parameterized by $\boldsymbol{\alpha}$, see (\ref{eq:thetaDirichlet}), because it is conjugated with the categorical distribution, thus providing an algebraic convenience for model inference

\begin{equation}\label{eq:thetaDirichlet}
\boldsymbol{\theta}_{d} \sim \mathrm{Dir}(K, \boldsymbol{\alpha})
\end{equation}
where $\boldsymbol{\alpha} = [\alpha_{1}, \dots, \alpha_{K}]$ is the vector-valued hyperparameter.

\subsubsection{Driving behavior} A specific driving style can generate different driving behaviors corresponding to the common factors $\{\mathbf{f}^{(\ell)}\}$ retrieved from Section II-A-2. We use a generative probabilistic model to formulate the relationship between a specific driving style and these common factors. The challenge is that the value of these factors falls into a continuous measure space, indicating that theoretically, the probability of any two behavior fragments being the identical value of the factors is zero. To solve this problem, we discretize each common factor into $M$ parts statistically in their measure space (see Section \ref{subsection: Data Extraction and Preprocessing}) using the techniques proposed in \cite{wang2018driving}. Thus, we can get a word-like set $\mathcal{W} = \{w_{i}\}_{i=1}^{M^m}$ of the discretized common factors, each element $w_{i}$ (called \textit{driving word}) represents the basic components of driving behavior. That is, each driving behavior fragment always tags one corresponding driving word from $\mathcal{W}$, making driving behavior characteristics shareable \textit{within} and \textit{between} human drivers. In such a way, we can represent driver $d$'s $j$-th behavior fragment using a driving word $w_{dj} \in \mathcal{W}$ and the driving behavior over time using a sequence of driving words, $\mathbf{w}_d=\left\{ w_{d1}, \dots, w_{dN}\right\}$. The generative probabilistic model for driving behavior and driving words allows us to evaluate the likelihood probability of belonging to a specific driving style for a given driving behavior. More specifically, 
for driver $d$, we use a categorical distribution parameterized by $\phi_{z_d}$ to formulate the relationship of a given driving style (e.g., $z_{d} = k$) and driving word $w_{d}$

\begin{equation}
\label{eq:driving behavior}
w_{d}\sim \mathrm{Cat}(M^m, \boldsymbol{\phi}_{k})
\end{equation}
with $\boldsymbol{\phi}_{k} = \left [\phi_{k1},  \dots, \phi_{ki},\dots, \phi_{kM^m} \right ]$ and $\sum_{i=1}^{M^m}\phi_{ki}=1$, where $\phi_{ki}$ represents the probability of assigning the driving word $w_{d}$ to driving style $k$. For a given sequence of driving behavior fragments of driver $d$, the count of each word-like element can be sampled from the multinomial distribution

\begin{equation}
\label{eq:pidMulti}
\boldsymbol{\pi}_{d} \sim \mathrm{Multi} ( N, \boldsymbol{\phi}_{k})
\end{equation}
with $\boldsymbol{\pi}_{d} = \left[ \pi_{d1}, \dots, \pi_{di}, \dots, \pi_{dM^{m}}\right]$ a vector-valued random variable, where $\pi_{di}$ is the count of $w_{i}$ in all driving behavior fragments. 

Usually, an aggressive driver would behave more aggressively than a calm driver. In other words, the aggressive driving style should probabilistically have a higher proportion for an aggressive driver than a calm one. To capture such mechanism, given a specific driving style $k$, we applied a Dirichlet distribution parameterized by $\boldsymbol{\beta}$ to governing the parameter $\boldsymbol{\phi}_{k}$,

\begin{equation}
\label{eq:phiDirchlet}
\boldsymbol{\phi}_k \sim \operatorname{Dir}(M^m, \boldsymbol{\beta}) 
\end{equation}
 where  $\boldsymbol{\beta} = \left[ \beta_1, \dots, \beta_{M^m}\right]$ is a vector-valued hyperparameter governing the distribution of aggressiveness for human drivers. 

\begin{figure}[t]
    \centering
    \includegraphics[width=0.8\linewidth]{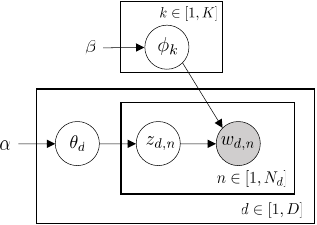}
    \caption{Graphical illustration of our proposed hierarchical latent driver model.}
    \label{fig: LDA}
\end{figure}

\subsubsection{Hierarchical Latent Driving Behavior Model}
Combining the Dirichlet-Category (Eqs. (\ref{eq:thetaDirichlet})-(\ref{eq:zdCat}) \& (\ref{eq:phiDirchlet})-(\ref{eq:driving behavior})) and Dirichlet-Multinomial (Eqs. (\ref{eq:thetaDirichlet})-(\ref{eq:cdMulti}) \& (\ref{eq:phiDirchlet})-(\ref{eq:pidMulti})) conjugations allows us to get driving behavior (i.e., behavior fragment) and latent shareable driving styles integrated into one single hierarchical latent model, as shown in Fig. \ref{fig: LDA}. More specifically, for a single \textit{driving fragment}, the Dirichlet-Category conjugation ($\boldsymbol{\alpha} \to \boldsymbol{\theta}_d \to z_d$) describes the embedding mechanism of a latent driving style $z$, and the categorical distribution $\boldsymbol{\phi}_{z_d} \to w_d$ formulates the generation mechanism of a driving behavior fragment with the generated driving style $z_d$. For a single \textit{human driver}, the Dirichlet-Multinomial conjugation ($\boldsymbol{\alpha} \to \boldsymbol{\theta}_d \to \mathbf{c}_d$) models the mixture proportion of given specific driving styles inherent in his/her minds, and the multinomial distribution $\boldsymbol{\phi}_k \to \boldsymbol{\pi}_d$ formulates the generation mechanism of the driving behavior counts with the generated driving styles $\mathbf{c}_d$. For a single specific driving style, the Dirichlet-Category conjugation ($\boldsymbol{\beta} \to \boldsymbol{\phi}_k \to w$) models the relationship between each specific driving style $k$ and driving words, which is shareable among all drivers. The Dirichlet-Category conjugacy and the Dirichlet-Multinomial conjugacy make this hierarchical latent driving behavior model computational and algebraic convergent. 

\subsection{Model Learning}

\begin{table*}[t]
\centering
\begin{minipage}{\textwidth}
\hrule
\begin{align}\label{eq: joint probability distribution}
    p(\mathbf{w}, \mathbf{z}, \boldsymbol{\Theta}, \boldsymbol{\Phi} \mid \boldsymbol{\alpha}, \boldsymbol{\beta})
    &=\underbrace{p(\mathbf{w} \mid \boldsymbol{\Phi}, \mathbf{z}) p(\boldsymbol{\Phi} \mid \boldsymbol{\beta})}
    _{\text {driving behavior generation}} \times  \underbrace{p(\mathbf{z} \mid \boldsymbol{\Theta}) p(\boldsymbol{\Theta} \mid \boldsymbol{\alpha})}
    _{\text { driving style generation}} \\
    & = \prod_{k=1}^{K} p(\boldsymbol{\phi}_t \mid \boldsymbol{\beta})  \prod_{d=1}^D \prod_{n=1}^{N_d} p\left(w_{d, n} \mid \boldsymbol{\phi}_{k=z_{d, n}}\right)  p\left(z_{d, n} \mid \boldsymbol{\Theta}\right) 
    p\left(\boldsymbol{\theta}_d \mid \boldsymbol{\alpha}\right) 
\end{align}
\medskip
\hrule
\end{minipage}
\end{table*}

According to the model structure illustrated in Fig~\ref{fig: LDA}, the joint distribution of all driving behavior fragments, driving styles, and their related (hyper)-parameters can be factorized as (\ref{eq: joint probability distribution}) (see next page) with $\boldsymbol{\Theta} = \{\boldsymbol{\theta}_1, \dots, \boldsymbol{\theta}_D \}$, where $D$ is the number of drivers and $\boldsymbol{\Phi} = \{\boldsymbol{\phi}_1, \dots, \boldsymbol{\phi}_K \}$.
The parameters $\boldsymbol{\alpha}$ and $\boldsymbol{\beta}$ are inferred using the collapsed Gibbs sampling based on the posterior distribution inference of latent driving styles $p(\mathbf{z} | \mathbf{w}, \boldsymbol{\alpha}, \boldsymbol{\beta})$. First, the factorization indicates that latent variable $\boldsymbol{\Theta}$ and $\boldsymbol{\Phi}$ can be integrated out separately, obtaining the joint distribution of driving words and driving styles (see Appendix \ref{sec:jointdistribution}),

\begin{equation}\label{Equation: joint distribution of behavior and style}
     p(\mathbf{z}, \mathbf{w} | \boldsymbol{\alpha}, \boldsymbol{\beta})
     = \prod_{k=1}^K \frac{\Delta\left(\mathbf{n}_k+\boldsymbol{\beta}\right)}{\Delta(\boldsymbol{\beta})} \prod_{d=1}^D \frac{\Delta\left(\mathbf{n}_d+\boldsymbol{\alpha}\right)}{\Delta(\boldsymbol{\alpha})}
\end{equation}
where $\Delta(\cdot)$ denotes the Dirichlet operator, $\mathbf{n}_{k} = \left\{ n_{z=k}^{i} \right \}_{i=1}^{M^m}$ is the counts of each driving word $w_i$ when driving style being $k$, $\mathbf{n}_{d} = \left\{n_d^{z=k}\right \}_{k=1}^{K}$ is the counts of each driving style shareable driving style $k$ for driver $d$.

Then, for driving style $z_i$ corresponding to the $i$-th driving behavior fragment, we get the fully conditional probability for $z_i =k$ as (see Appendix \ref{sec: posterior distribution})

\begin{equation} \label{Equation: posterior probability of driving style i}
    p\left(z_i | \mathbf{z}_{\neg i}, \mathbf{w}, \boldsymbol{\alpha}, \boldsymbol{\beta}\right) 
    \propto \frac{n_k^i+\beta_i}{\sum_{j=1}^M\left(n_k^j+\beta_j\right)}  \frac{n_d^k+\alpha_k}{\sum_{k=1}^K\left(n_d^k+\alpha_k\right)},   
\end{equation}
where $\mathbf{z}_{\neg i} $ denotes all driving styles except for $z_{i}$, $\mathbf{w}$ are the observed driving behavior fragments, and $\mathbf{w}_{\neg i} $ denotes the all other driving behavior fragments excluding the current one $w_i$.
Equation (\ref{Equation: posterior probability of driving style i}) is used in Gibbs sampling to approximate the posterior distribution of the driving styles among the test drivers. 

\subsection{Evaluation Criteria}
It is a fact that a good driving behavior model should assign high probabilities to frequent driving behavior fragments and low probabilities to randomly generated and meaningless ones. The best model should assign the lowest perplexity to the dataset. We use two metrics of perplexity and Shannon entropy \cite{blei2003latent,qi2015appropriate,chen2021exploring} to quantify our proposed model performance.
\subsubsection{Perplexity} The perplexity describes the \emph{uncertainty} of driving dataset using the inverse probability as 

\begin{equation}
\label{eq: perplexity}
    \eta =\left(\frac{1}{\prod_{d=1}^{D} p(\mathbf{w}_d )}\right)^{\frac{1}{N}} 
         =\exp{\left(-\frac{\sum _{d=1}^D \log p(\mathbf{w}_d)}{N} \right)}
\end{equation}
where $\log p(\mathbf{w}_d)$ is the log-likelihood probability of all the word-like sequential driving behavior for driver $d$, and $N = \sum _{d=1}^D N_d $. We normalize the probability of the dataset by $N$ in (\ref{eq: perplexity}) to make the perplexity independent of dataset size, thus providing a fair evaluation for individual driving behavior fragments. However, a denser common factor discretization (see Section II-B-2) results in a larger size $M^m$ of the driving word set $\mathcal{W}$, thus introducing more uncertainty for individuals' behavior fragments. In other words, a large value of $M$ will increase the perplexity. To make the metric independent of the size of the driving word set, we normalize the perplexity as $\bar{\eta} = \frac{\eta}{|\mathcal{W}|}$, $|\mathcal{W}|$ is the set size.

\subsubsection{Shannon Entropy} Each driving style corresponds to one parameter variable $\boldsymbol{\phi}_k$ governing a categorical distribution over driving behavior. The variable $\boldsymbol{\phi}_k$ describes the probability of the driving style $k$ assigning to each driving behavior. For example, a driving style with a high probability of exhibiting violent acceleration driving behavior is likely to be aggressive. Therefore, the lower the entropy of the driving style, the clearer semantics of the driving style. Thus, we quantify the semantic clarity of learned driving styles using averaged Shannon entropy as

\begin{equation}
    H_{\text{styles}}= \frac{1}{K}\sum_{k=1}^{K} H_k = -\frac{1}{K}\sum_{k=1}^{K} \sum_{i=1}^{M^m} \phi_{ki} \log \phi_{ki}
\end{equation}
where $H_k$ is the Shannon entropy of a single driving style $k$. 

\section{Data Collection and Experimental Setting}
\subsection{Data Collection}

\begin{figure}
    \centering
    \includegraphics[width=\linewidth]{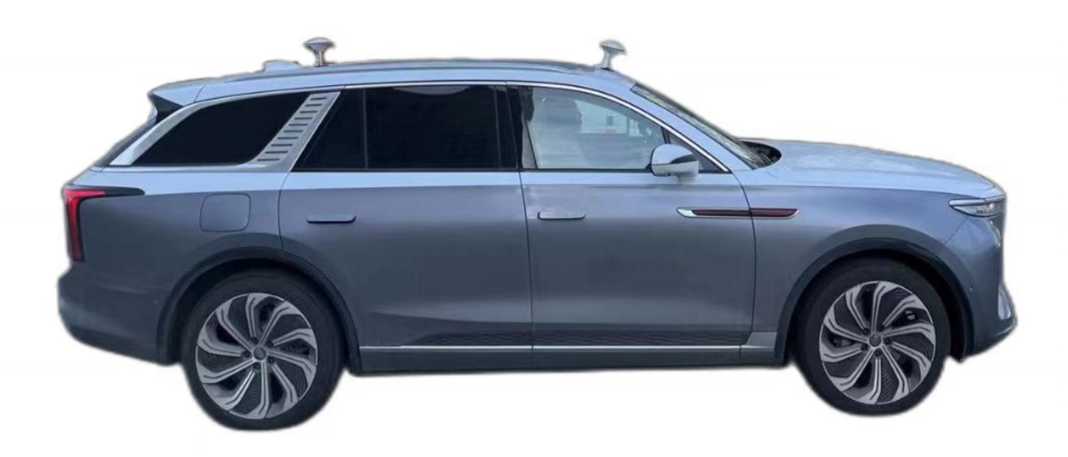}
    \caption{The testing vehicle Hongqi E-HS9 for data collection.}
    \label{fig: Equipment}
\end{figure}

We validate our proposed method in naturalistic traffic settings with a Hongqi E-HS9 testing vehicle equipped with a data-acquisition system (see Fig.~\ref{fig: Equipment}), consisting of a vehicle CAN bus, an integrated navigation system, and one front-view camera. The vehicle CAN information provides speed, acceleration, steering angle, etc. The integrated navigation system obtains vehicle positioning and orientation information. The camera collects traffic information, such as the distance to the leading vehicle, traffic lights, and lane lines.

\begin{figure}[t]
    \centering
    \includegraphics[width=\linewidth]{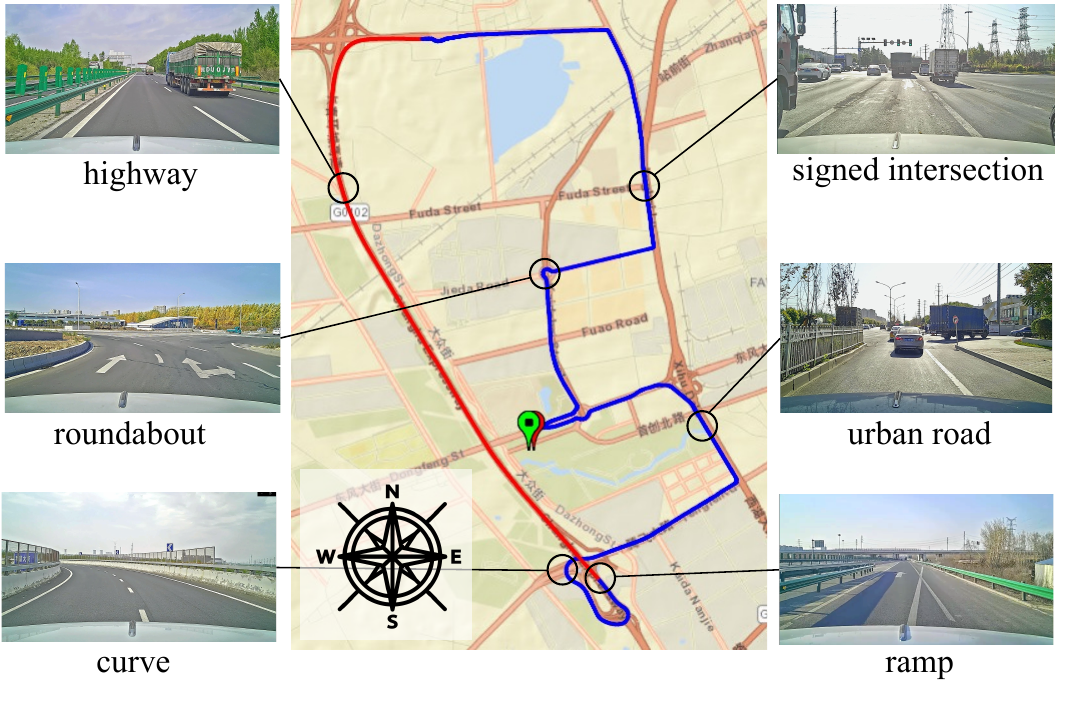}
    \caption{Selected driving routes with highways (red line) and urban (blue line) settings in Changchun, China.}
    \label{fig: Driving routes}
\end{figure}

All real-time acquisitions are synchronized and recorded using CANoe at $100$ Hz. The CAN bus load rate is less than $30\%$, ensuring data transmission stability. We use longitudinal velocity, acceleration, and yaw rate to represent driving behavior because these signals are available in most production cars, making it tractable to adapt the algorithm to other vehicles. We select the driving routes in Changchun, China, with $14.4$ km of urban roads and $7.6$ km of highways, covering various driving scenarios, as shown in Fig.~\ref{fig: Driving routes}. 

\begin{table}[t]
\centering
\caption{The Summary of Driver Participants}
\label{tab: Statistical summary of tested drivers}
\begin{tabular}{ll|ll|ll}
\hline\hline
\multicolumn{2}{l|}{Age} & \multicolumn{2}{l|}{Driver experience} & \multicolumn{2}{l}{Occupation} \\
\hline
Range      & Number      & Range             & Number             & Occupation        & Number     \\
\hline
$21-25$      & $3$           & $1-5$               & $12$                 & bus driver        & $38$         \\
$26-30$      & $8$           & $6-10$              & $20$                 & taxi driver       & $10$         \\
$31-35$      & $14$          & $11-15$             & $21$                 & ride-hailing      & 9          \\
$36-40$      & $21$          & $16-20$             & $20$                 & self-employed     & $9$          \\
$41-45$      & $18$          & $21-25$             & $12$                 & engineer          & $4$          \\
$46-50$      & $16$          & $26-30$             & $8$                  & worker            & $4$          \\
$51-60$      & $17$          & $31-35$             & $7$                  & teacher           & $3$          \\
$56-60$      & $3$           &-                    &-                 & freelancer        & $16$         \\- 
           &-              &-                    & -                    & others            & $7$         \\
\hline\hline
\end{tabular}
\end{table}

\subsection{Driver Participants and Test Procedure}
\label{subsection: Driver Participants}
Human personalities, such as age, gender, occupation, and driving experience, can influence driving styles. We recruited $100$ drivers ($83$ males and $17$ females) with diverse personalities, as shown in Table \ref{tab: Statistical summary of tested drivers}. The test procedure is as follows.

Each driver is asked to take about $60$ minutes of participation, including a warm-up, natural driving, and filling out a questionnaire. To get the participants used to the vehicle's operation, each participant will take a $20$-minute warm-up session for driving the testing car according to their habits. We will explain safety issues to the driver and clarify doubts. The driver will also preview the navigation route. 

After the warm-up session, the participant will drive naturally according to their driving style by following the selected navigation route. Slight speeding is allowed because we found that some drivers often exceed the speed limit in their daily driving without traffic monitoring. This reflects the participant's driving style, albeit irrational. Once the data-collection session is triggered, our data-acquisition system automatically records the participant's driving operations. 

We also conducted a \textit{subjective} evaluation of driving style for each participant, including self-report and experts' scoring. More specifically, we asked each driver participant to complete a well-prepared driving style questionnaire, providing their age, gender, occupation, and driving experience, to report their own driving styles. The driving style questionnaire is designed by following the Likert scale \cite{joshi2015likert,padilla2020adaptation}, an easy-to-answer questionnaire paradigm, considering safety, risk-taking, and stimulation-seeking, as shown in Appendix \ref{app:likertscale}. The driver participants do not have to answer exactly what type of driving style they belong to but indicate their aggressiveness level. In addition, we also invited an expert sitting at the co-pilot seat of the testing vehicle to subjectively score the participant's driving style.

\subsection{Data Preprocessing}
\label{subsection: Data Extraction and Preprocessing}

\begin{table}[t]
\centering
    \caption{Correlation (i.e., Factor Loading) Between Variables and Common Factors}
    \label{tab: Factor loading}
\begin{tabular}{lrrr}
\hline\hline
                               & \multicolumn{3}{c}{Factor loading}                                                                       \\   \cline{2-4}
\multirow{-2}{*}{$\mathbf{y}$} & \multicolumn{1}{c}{$\mathbf{f}^{\mathrm{lat}}$} & \multicolumn{1}{c}{$\mathbf{f}^{\mathrm{acc}}$} & \multicolumn{1}{c}{$\mathbf{f}^{\mathrm{speed}}$} \\
 \hline
$v^{\mathrm{\max}}$ & $-0.014$   & $0.157$  & \cellcolor{gray!30}$0.978$\\
$v^{\mathrm{avg}}$  & $-0.156$  & $-0.146$  & \cellcolor{gray!30}$0.965$ \\
$v^{\mathrm{std}}$ & $0.069$  &  \cellcolor{gray!30}$0.858$ & $0.151$ \\
$a_{x}^{\mathrm{\max}}$  & $0.076$ & \cellcolor{gray!30}$0.938$ & $-0.063$ \\
$a_{x}^{\mathrm{avg}}$  & $0.057$ & \cellcolor{gray!30}$0.934$ & $-0.041$ \\
$a_{x}^{\mathrm{std}}$ & $0.090$  &  \cellcolor{gray!30}$0.900$ & $-0.039$  \\
$\psi^{\mathrm{\max}}$ & \cellcolor{gray!30}$0.944$ & $0.137$ & $-0.175$ \\
$\psi^{\mathrm{avg}}$  &  \cellcolor{gray!30}$0.914$  & $-0.055$ & $-0.154$  \\
$\psi^{\mathrm{std}}$   & \cellcolor{gray!30}$0.904$ & $0.176$  & $-0.159$ \\
$a_{y}^{\mathrm{\max}}$  & \cellcolor{gray!30}$0.962$  & $0.108$ & $-0.011$ \\
$a_{y}^{\mathrm{avg}}$  & \cellcolor{gray!30}$0.915$  & $-0.055$ & $-0.006$  \\
$a_{y}^{\mathrm{std}}$  & \cellcolor{gray!30}$0.942$ & $0.143$ & $-0.011$\\
 \hline \hline
\end{tabular}
\end{table}
\begin{figure}[!ht]
    \centering
    \includegraphics[scale=0.95]{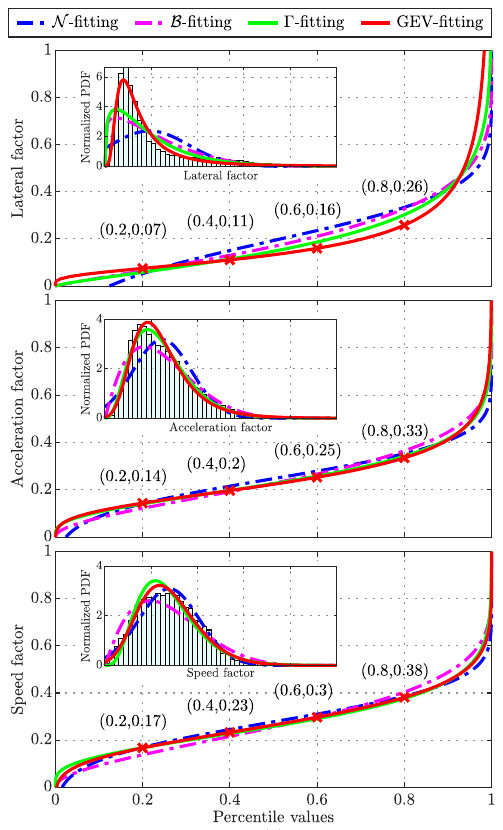}
    \caption{The statistical fitting results using four distributions, i.e., normal distribution ($\mathcal{N}$), Beta distribution ($\mathcal{B}$), Gamma distribution ($\Gamma$) and generalized extreme value distribution (GEV) and  the segmentation of three factors by probability}
	\label{fig: Factor statistical fitting}
\end{figure}

\subsubsection{Dimension Reduction}
The driving behavior analysis is based on vehicle CAN signals, including speed $v$, longitudinal acceleration $a_x$, lateral acceleration $a_y$, and yaw rate $\psi$. We set $\tau = 10$s and compute each driving behavior fragment's statistics (mean, maximum, standard deviation) to feature the driving style (see Table \ref{tab: Factor loading}).
We then reduce the dimension of $\mathbf{y}$ using the approach introduced in Section II-A-2) to obtain the common factors. Finally, we obtain three explainable common factors according to Kaiser's criterion \cite{kaiser1960application}. The value close to $1$ indicates a strong positive correlation between $\mathbf{y}$ and associated common factors. The first factor obtains a high value close to $1$ for all lateral motion variables, named lateral factor, $\mathbf{f}^{\mathrm{lat}}$. The second factor obtains a high value close to $1$ for the acceleration-related variables, named acceleration factors, $\mathbf{f}^{\mathrm{acc}}$. The third factor obtains a high value close to $1$ for the speed-related variables, named the speed factors, $\mathbf{f}^{\mathrm{speed}}$. In other words, the driving style representation relies on the steering-, acceleration-, and speed-related features.

\begin{figure}[t]
 \centering
    \includegraphics[width=0.75\linewidth]{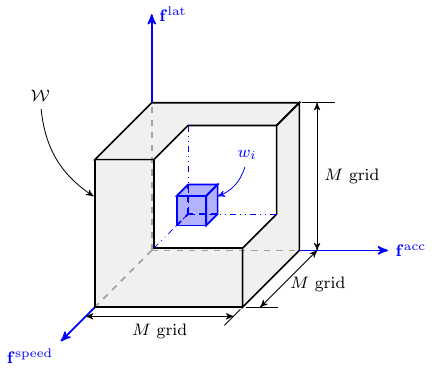}
    \caption{Illustration of the set $\mathcal{W}$ of the driving word with a size of $M^{3}$ corresponding to the three common factors.}
    \label{fig: DiscreteCommonFactor}
\end{figure}

\begin{figure}
    \centering
    \includegraphics[scale=1]{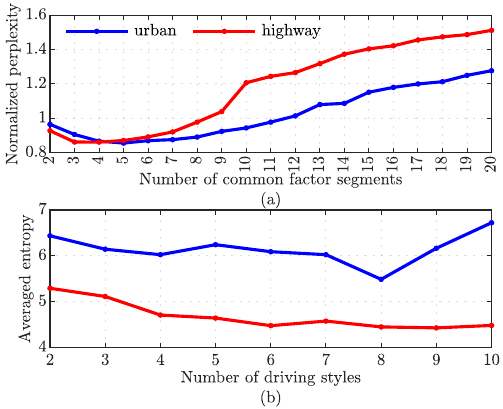}
    \caption{Results of perplexity and entropy with different numbers of word segments and topics in urban and highways.}
    \label{fig: perplexity_entropy}
\end{figure}

\subsubsection{Word-like Set of Driving Behavior}
Based on the common factors learned with dimension reduction, we independently grid each common factor into $M$ parts using the probabilistic approach introduced in Section II-B-2). Note that a joint distribution could theoretically formulate the distribution of the three common factors. However, the great distinction in the three common factors' distribution (see  Fig. \ref{fig: Factor statistical fitting}) makes it hard to achieve this jointly. Therefore, we fit the  lateral factor ($\mathbf{f}^{\mathrm{lat}}$), acceleration factor ($\mathbf{f}^{\mathrm{acc}}$), speed factor ($\mathbf{f}^{\mathrm{speed}}$) independently with the normal distribution ($\mathcal{N}$), Beta distribution ($\mathcal{B}$), Gamma distribution ($\Gamma$), and generalized extreme value distribution (GEV). The GEV distribution achieves the best-fitting performance for the lateral and acceleration factors. The GEV and $\mathcal{N}$-distribution obtains the best fitting performance for the speed factor. In the distributions of the three common factors, the large factor values are distributed in the tail, representing aggressive driving. Thus, we use the percentile values of the three common factors based on the GEV-fitting results due to its advantage in fitting heavy-tailed data\cite{hosking1985estimation}.  Finally, we obtain the set of driving words with a size of $M^{3}$ through the permutations of the discretized common factors, as illustrated in Fig. \ref{fig: DiscreteCommonFactor}. 

\subsection{Hyperparameters Setting}
\subsubsection{Number of Common Factor Segments $M$} 
\label{subsubsection: Driving Word Types Determination}
Determining the number of segments $M$ for each common factor is essential. We set different $M$ in a range from $2$ to $20$ to train our model separately. Fig.~\ref{fig: perplexity_entropy}(a)  shows the testing results and indicates that $M=5$ in urban and $M=4$ in highways make the lowest perplexity of driving word distribution. 

\subsubsection{Number of Shareable Driving Styles $K$}
\label{section: Driving Topic Number Determination}
The other parameter to be set is the number of shareable driving styles $K$. We utilize entropy to determine the optimal value of $K$ as similar to \cite{chen2021exploring} while considering the interpretation of driving styles. Although Fig.~\ref{fig: perplexity_entropy}(b) implies that $K\ge 6$ might make the lowest entropy of driving styles, some driving styles are uniform probability distributions over driving words and are meaningless for driving behavior analysis. Therefore, considering the application of driving style to intelligent vehicles and eaning of driving styles, we set the number of shareable driving styles $K=3$ (i.e., aggressive, calm, and normal) both in urban and highways based on the categories of commonly used \cite{samani2022assessing,martinez2017driving,xu2015establishing,yang2019time}.

\begin{figure}[t]
    \centering
    \includegraphics[scale=1]{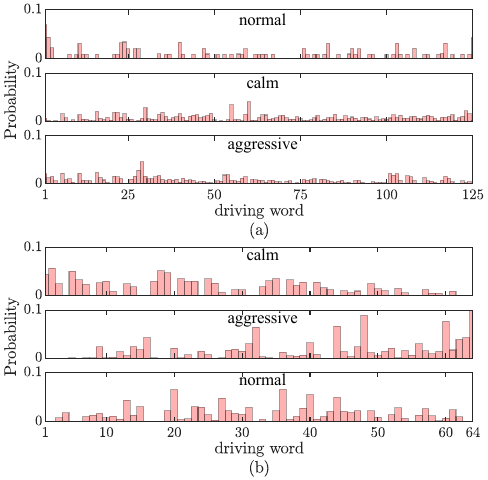}
    \caption{The distribution of driving words $w_{i}$ for each shareable driving style in the settings of (a) urban and (b) highways.}
    \label{fig: Different Topic distributions over driving words}
\end{figure}

\section{Experiment Results and Analysis}
\label{sec:experimentresults}

\subsection{Subjective-Objective Consistency Verification}

\begin{table}[t]
\centering
\caption{Statistics of Driving Styles with Mean, Standard Deviation (sd), 25th Percentile ($Q^{25\mathrm{th}}$), Median ($Q^{\mathrm{med}}$), and 75th Percentile ($Q^{75\mathrm{th}}$).}
\vspace{1ex}
\label{tab: Static performance of different driving topics}
\begin{tabular}{lcccc|ccc}
\hline\hline
                          &                         & \multicolumn{3}{c}{Drving styles in urban}                                                           & \multicolumn{3}{c}{Driving styles in highway}                                                           \\ \cline{3-8} 
\multirow{-2}{*}{} & \multirow{-2}{*}{Index} & normal                             & calm                             & aggr.                             & calm                             & aggr.                              & normal                              \\ 
\hline
                          & Mean                    & \cellcolor{blue!30}$54.60$ & \cellcolor{green!30}$43.24$ & \cellcolor{red!30}$59.00$ & \cellcolor{green!30}$84.05$ & \cellcolor{red!30}$106.4$ & \cellcolor{blue!30}$96.72$  \\
                          & SD                      & \cellcolor{blue!30}$18.24$ & \cellcolor{green!30}$12.64$ & \cellcolor{red!30}$18.30$ & \cellcolor{green!30}$12.84$ & \cellcolor{red!30}$19.92$  & \cellcolor{blue!30}$19.75$  \\
                          & $Q^{25\mathrm{th}}$                      & \cellcolor{blue!30}$40.44$ & \cellcolor{green!30}$32.96$ & \cellcolor{red!30}$46.12$ & \cellcolor{green!30}$73.97$ & \cellcolor{red!30}$92.58$  & \cellcolor{blue!30}$78.07$  \\
                          & $Q^{\mathrm{med}}$                      & \cellcolor{blue!30}$54.39$ & \cellcolor{green!30}$41.96$ & \cellcolor{red!30}$60.45$ & \cellcolor{green!30}$84.77$ & \cellcolor{red!30}$107.8$ & \cellcolor{blue!30}$101.9$ \\
\multirow{-5}{*}{$v$}       & $Q^{75\mathrm{th}}$                      & \cellcolor{blue!30}$68.46$ & \cellcolor{green!30}$54.17$ & \cellcolor{red!30}$71.32$ & \cellcolor{green!30}$93.95$ & \cellcolor{red!30}$119.0$ & \cellcolor{blue!30}$111.2$ \\ 
\hline
                          & Mean                    & \cellcolor{blue!30}$0.59$  & \cellcolor{green!30}$0.38$  & \cellcolor{red!30}$0.80$  & \cellcolor{green!30}$0.18$  & \cellcolor{red!30}$0.47$   & \cellcolor{blue!30}$0.32$   \\
                          & SD                      & \cellcolor{blue!30}$0.55$  & \cellcolor{green!30}$0.35$  & \cellcolor{red!30}$0.65$  & \cellcolor{green!30}$0.17$  & \cellcolor{red!30}$0.44$   & \cellcolor{blue!30}$0.32$   \\
                          & $Q^{25\mathrm{th}}$                      & \cellcolor{blue!30}$0.16$  & \cellcolor{green!30}$0.12$  & \cellcolor{red!30}$0.27$  & \cellcolor{green!30}$0.04$  & \cellcolor{red!30}$0.14$   & \cellcolor{blue!30}$0.09$   \\
                          & $Q^{\mathrm{med}}$                     & \cellcolor{blue!30}$0.44$  & \cellcolor{green!30}$0.30$  & \cellcolor{red!30}$0.65$  & \cellcolor{green!30}$0.13$  & \cellcolor{red!30}$0.36$   & \cellcolor{blue!30}$0.23$   \\
\multirow{-5}{*}{$a_x$}       & $Q^{75\mathrm{th}}$                     & \cellcolor{blue!30}$0.86$  & \cellcolor{green!30}$0.53$  & \cellcolor{red!30}$1.20$  & \cellcolor{green!30}$0.25$  & \cellcolor{red!30}$0.66$   & \cellcolor{blue!30}$0.46$   \\ 
\hline
                          & Mean                    & \cellcolor{blue!30}$1.55$  & \cellcolor{green!30}$1.52$  & \cellcolor{red!30}$1.77$  & \cellcolor{green!30}$0.41$  & \cellcolor{red!30}$0.58$   & \cellcolor{blue!30}$0.50$   \\
                          & SD                      & \cellcolor{blue!30}$2.89$  & \cellcolor{green!30}$2.81$  & \cellcolor{red!30}$2.97$  & \cellcolor{green!30}$0.67$  & \cellcolor{red!30}$1.03$   & \cellcolor{blue!30}$0.81$   \\
                          & $Q^{25\mathrm{th}}$                      & \cellcolor{blue!30}$0.15$  & \cellcolor{green!30}$0.14$  & \cellcolor{red!30}$0.21$  & \cellcolor{green!30}$0.09$  & \cellcolor{red!30}$0.12$   & \cellcolor{blue!30}$0.09$   \\
                          & $Q^{\mathrm{med}}$                      & \cellcolor{blue!30}$0.45$  & \cellcolor{green!30}$0.40$  & \cellcolor{red!30}$0.60$  & \cellcolor{green!30}$0.24$  & \cellcolor{red!30}$0.33$   & \cellcolor{blue!30}$0.25$   \\
\multirow{-5}{*}{$\phi$}     & $Q^{75\mathrm{th}}$                      & \cellcolor{blue!30}$1.26$  & \cellcolor{green!30}$1.18$  & \cellcolor{red!30}$1.70$  & \cellcolor{green!30}$0.45$  & \cellcolor{red!30}$0.60$   & \cellcolor{blue!30}$0.50$   \\ 
\hline
                          & Mean                    & \cellcolor{blue!30}$0.32$  & \cellcolor{green!30}$0.27$  & \cellcolor{red!30}$0.41$  & \cellcolor{green!30}$0.15$  & \cellcolor{red!30}$0.25$   & \cellcolor{blue!30}$0.18$   \\
                          & SD                      & \cellcolor{blue!30}$0.47$  & \cellcolor{green!30}$0.42$  & \cellcolor{red!30}$0.56$  & \cellcolor{green!30}$0.19$  & \cellcolor{red!30}$0.32$   & \cellcolor{blue!30}$0.22$   \\
                          & $Q^{25\mathrm{th}}$                      & \cellcolor{blue!30}$0.05$  & \cellcolor{green!30}$0.04$  & \cellcolor{red!30}$0.08$  & \cellcolor{green!30}$0.04$  & \cellcolor{red!30}$0.06$   & \cellcolor{blue!30}$0.04$   \\
                          & $Q^{\mathrm{med}}$                      & \cellcolor{blue!30}$0.14$  & \cellcolor{green!30}$0.13$  & \cellcolor{red!30}$0.20$  & \cellcolor{green!30}$0.11$  & \cellcolor{red!30}$0.15$   & \cellcolor{blue!30}$0.12$   \\
\multirow{-5}{*}{$a_y$}      & $Q^{75\mathrm{th}}$                      & \cellcolor{blue!30}$0.33$  & \cellcolor{green!30}$0.27$  & \cellcolor{red!30}$0.48$  & \cellcolor{green!30}$0.20$  & \cellcolor{red!30}$0.31$   & \cellcolor{blue!30}$0.24$   \\ 
\hline\hline
\end{tabular}
\end{table}

We assign semantics to shareable driving styles according to their driving data statistics.  Fig. \ref{fig: Different Topic distributions over driving words} shows the distribution of driving words for latent driving styles. Recall that the numbers of driving words for urban and highway scenarios differ, according to the settings of $M$ in Section III-D-1). We also list each feature's statistics (means, standard deviation, and percentiles) for all latent driving styles in Table \ref{tab: Static performance of different driving topics} to assign semantics to them. The aggressive style obtains the highest score in all statistic indexes, and the calm style corresponds to the lowest score. The normal style's index values are in the middle. And then, each driver's driving style can be recognized as a mixture proportion $\boldsymbol{\theta}$ of the three shareable driving styles in our proposed hierarchical latent driving  behavior model. In this way, we summarize the diverse personalized driving behavior of a driver into a low-dimensional latent presentation (i.e., embedding) \cite{shi2015evaluating}. The embedding could serve as a clear and quantitative description of driving styles.

To verify the performance of the proposed method (i.e., the objective results), we compare drivers' driving style mixtures with the subjective evaluation of driving style, $s_{\mathrm{sub}}$, which is a discrete value scored by experts and participant themselves.  However, the driving style mixtures are continuous, prohibiting the comparison. To make subjective-objective comparison tractable, we calculate each driver's objective aggressive score $s_{\mathrm{obj}}$ using their expected driving style

\begin{equation}
\label{Equation: the aggressive score}
    s_{\mathrm{obj}} = \sum^K_{k=1} \gamma_{k} \theta_{k}
\end{equation}
where $\theta_{k}$ is the proportion of the driving style $k$ for individual drivers, $\gamma_{k}$ is the corresponding weight. It should be noted that we reorder the driving styles according to their aggressiveness. That is, a more aggressive driving style corresponds to a larger value of $k$. Table \ref{tab: The threshold interval used to convert the score into the driving style} lists the thresholds used to convert these scores into discrete driving styles regarding aggressive levels, which makes it possible to use subjective results to evaluate our proposed model performance. We set $\gamma_k = k$ and find the optimal value by traversal.

\begin{table}[t]
\centering
\caption{The Thresholds for Converting the Objective Score ($s_{\mathrm{obj}}$) into the Aggressive Levels}
\label{tab: The threshold interval used to convert the score into the driving style}
\begin{tabular}{p{1cm}p{1cm}p{1cm}p{1cm}p{1cm}p{1cm}}
\hline\hline
     & \multicolumn{5}{c}{Aggressive level} \\
\cline{2-6}
     & $1$      & $2$         & $3$         & $4$         & $5$      \\
\hline
urban & $[1.00,1.01]$ & $[1.01,1.37]$ & $[1.37,1.99]$ & $[1.99,2.97]$ & $[2.97,3]$ \\
highway & $[1.00,1.03]$ & $[1.03,1.57]$ & $[1.57,2.19]$ & $[2.19,2.96]$ & $[2.96,3]$ \\
\hline\hline
\end{tabular}
\end{table}

\begin{figure}[t]
    \centering
    \includegraphics[width=\linewidth]{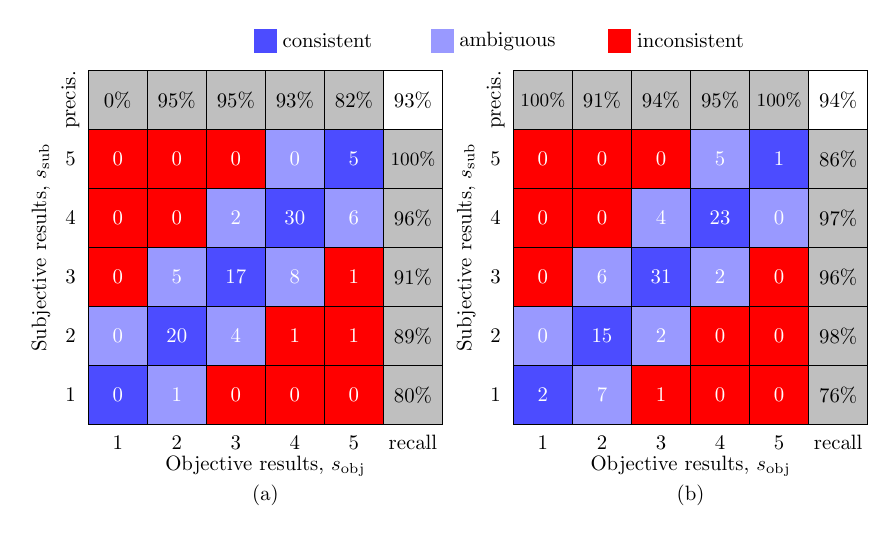}
    \caption{An illustration of the consistency weighted confusion matrix and performance in subjective-objective consistency verification in (a) urban and (b) highway settings.}
    \label{fig: An illustration of the confusion matrix for driving style classification}
\end{figure}

To compare our proposed model performance with the subjective evaluation of driving style, we proposed to use the subjective-objective \textit{consistency}, illustrated as a weighted confusion matrix (see Fig. \ref{fig: An illustration of the confusion matrix for driving style classification}), wherein the horizontal and vertical axis represent objective and subjective results, respectively. Thus, the differences between vertical and horizontal coordinates (i.e., the matrix elements that are not on the matrix diagonal) demonstrate the subjective-objective consistency of driving styles. A small value of the differences indicates a high subjective-objective consistency. For convenient understanding,  we named the results according to their coordinate differences: consistent when subjective-objective difference $=0$, ambiguous when subjective-objective difference $=1$, and inconsistent when subjective-objective difference $\ge 2$. This is intuitively reasonable since there does not exist an apparently-strict and fully-convincing boundaries to distinguish two adjacent subjective/objective driving styles. Further, to consider the contribution of the subjective-objective difference to the consistency performance, we assign  a consistency weight of $1.0$ to consistent results and  a consistency weight of $0$ to inconsistent results. For ambiguous results, the objective results are very close to the subjective results, so we assign them a consistency weight of $0.8$.  And then we compute the accuracy, precision, and recall according to the consistency weighted confusion matrix. Neither method identified driving style.

Fig. \ref{fig: An illustration of the confusion matrix for driving style classification} implies that for urban data, the general classification accuracy of our proposed method for five aggressive levels is $
93\%$. Notice that aggressive level $1$ is not recognized because there might exist a few extremely mild driver samples ($1\%$) and the distinction between aggressive levels $1$ and $2$ is not apparent. However, the recall of aggressive level $5$ reaches $100\%$ because the aggressive drivers' driving behavior is distinctive, such as high speed (or speeding), large throttle opening, and large steering angle. Similarly, aggressive level $4$ performs well in precision ($93\%$) and recall ($96\%$). For aggressive level $3$, our model gets desired performance in precision of $95\%$ and recall of $91\%$. It should be noticed that aggressive level $3$ might be easily identified as $2$ or $4$ because the boundary between adjacent styles is ambiguous. This is confirmed in Fig.~\ref{fig: An illustration of the confusion matrix for driving style classification} (a): $8/30$ samples of level $3$ are identified as $4$, and $5/30$ samples of level $3$ are identified as $2$. For highway data,  the general classification accuracy of our proposed model for the five aggressive levels is $94\%$.  The method performs well in all levels except aggressive level $1$. The precision of aggressive level $1$ reaches $100\%$ but with a low recall value ($76\%$). In conclusion, our proposed hierarchical latent driving behavior model performs well in subjective and objective consistency verification in urban and highway settings. In the next sections, we will show the illustrations of shareable driving styles \textit{within} and \textit{between} individuals and some important findings based on the proposed method.

\begin{figure*}[htp]
    \centering
    \includegraphics[]{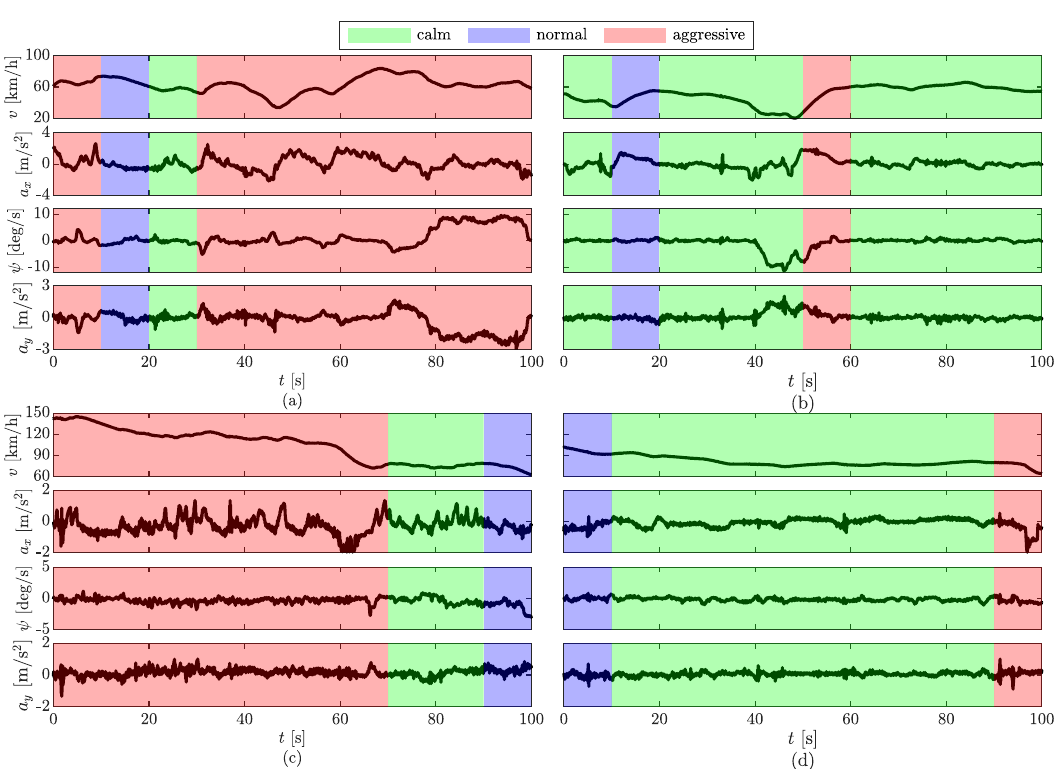}
    \caption{Examples of illustrating the learned driving styles in different environment settings: (a) An aggressive driver in urban, (b) a calm driver in urban, (c) an aggressive driver in highways, and (d) a calm driver in highways.}
    \label{fig: Examples of changes in driving style.}
\end{figure*}

\begin{figure}[htp]
    \centering
    \includegraphics[width=0.95\linewidth]{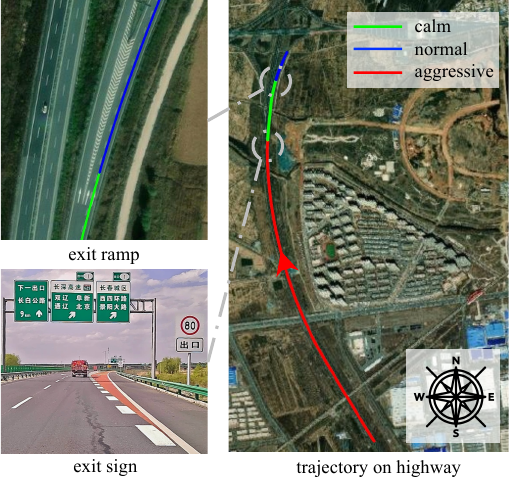}
    \caption{The driving trajectory of the examples shown in Fig.~\ref{fig: Examples of changes in driving style.} (c).}
    \label{fig: Trajectory of the aggressive driving on highways.}
\end{figure}
\subsection{Driving Style Distributions \textit{Within} Individuals}
\textbf{1) An aggressive driver prefers to behave more aggressively than a calm driver.} Fig.~\ref{fig: Examples of changes in driving style.}(a) and (b) show examples of changes in driving style for an aggressive driver and a calm driver when driving in urban. For the aggressive driver, the illustrated driving data changes drastically during aggressive driving. For instance, during $[30, 100]$s, the longitudinal and lateral acceleration (i.e., $a_x$, $a_y$)  changes drastically, so the driving style is aggressive. During $[0, 10]$s, speed $v$ exceeds the speed limit (60 km/h),  and $a_x$ changes, causing an aggressive driving style. In contrast to the aggressive driver, the driving data of the calm driver do not change drastically with small values. Specifically, during $[0, 10]$s and $[20, 50]$s, the speed is low, resulting in a calm driving style. During $[60, 100]$s, the driver keeps straight and maintains a constant speed, so the driving style is identified as calm. Fig.~\ref{fig: Examples of changes in driving style.}(c) and (d) show that the aggressive driver drives more aggressively than the calm driver when driving on highways.

\textbf{2) Driving styles are habit-driven but will still change with the environment.} As shown in Fig.~\ref{fig: Examples of changes in driving style.}(c), the aggressive driver drives aggressively from $0$ to $70$s but then turns calmer from $70$ to $100$s. The reason is that the driver saw the exit and speed limit signs (see Fig.~\ref{fig: Trajectory of the aggressive driving on highways.}) and kept a low speed from $70$ to $90$s, so the driving style turned calm from aggressive. From $90$ to $100$s, the driver turned onto the highway exit ramp, and the yaw rate $\phi$ changed abruptly, corresponding to the changes in driving style from calm to normal. 
\begin{figure*}[t]
    \centering
    \includegraphics[width=0.98\linewidth]{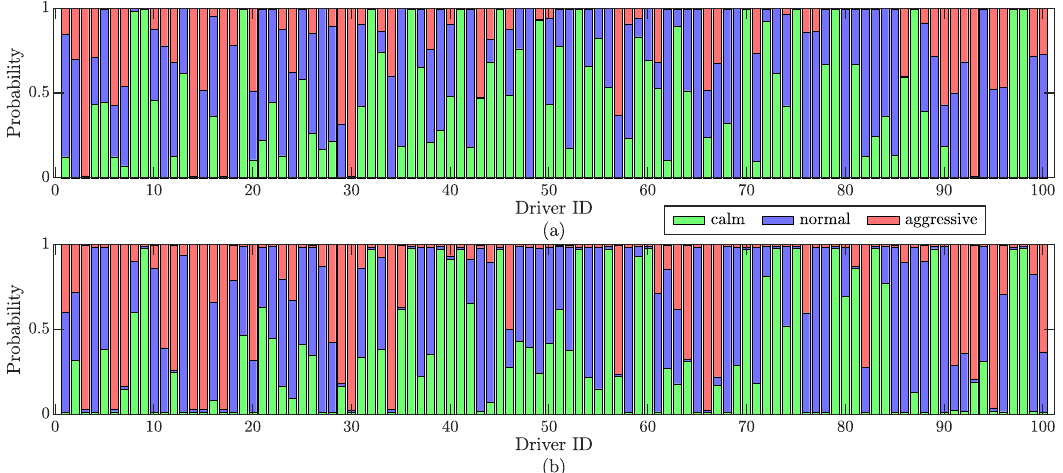}
    \caption{Shareable driving style mixtures of the 100 test drivers in (a) urban and (b) on highways.}
    \label{fig: Distribution of topic for 100 drivers}
\end{figure*}

\begin{figure*}[t]
    \centering
    \includegraphics[]{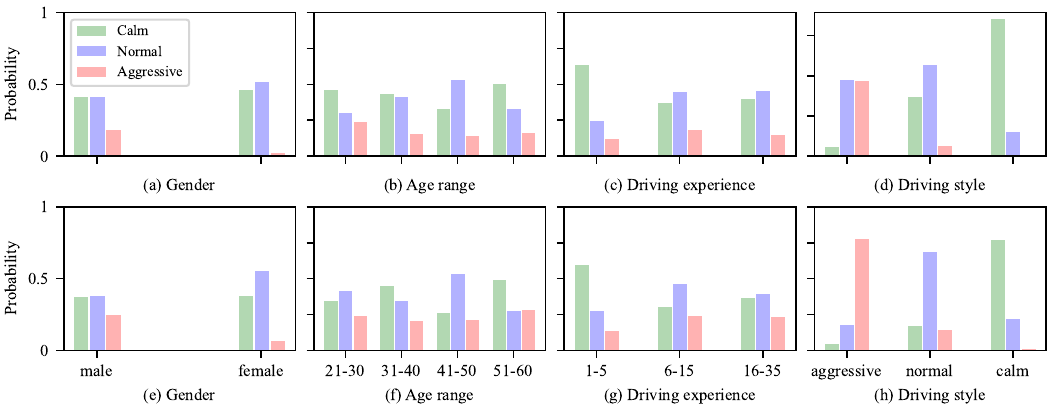}
    \caption{Driving topic distributions across drivers groups in urban (top row) and on highways (bottom row).}
    \label{fig: Driving topic distributions across different drivers}
\end{figure*}
\subsection{Driving Style Distributions \textit{Between} Individuals}

\textbf{3) Driving styles are shareable \textit{within} individuals.}  Fig.~\ref{fig: Examples of changes in driving style.}  and \ref{fig: Trajectory of the aggressive driving on highways.} reveal that the driver could exhibit the same style at different moments and places of driving. For instance, Fig.~\ref{fig: Examples of changes in driving style.} (a) shows that the aggressive driver driving aggressively both in $[0, 10]$s and $[30, 100]$s, which indicates driving styles are temporally shareable. Fig.~\ref{fig: Examples of changes in driving style.} (b) and (d) the calm driver could drive calmly both in urban and highways, which indicates driving styles are spatially shareable.

\textbf{1) Driving styles are shareable \textit{between} individuals.} 
Fig. ~\ref{fig: Distribution of topic for 100 drivers} shows the specific driving style mixtures of individual drivers in urban and highways. Although individual drivers drive in their way, they share the same three driving styles involved in driving behavior. As illustrated, the ratio of aggressive, normal, and calm driving styles across individual drivers differ. For instance, the aggressive style of driver \#$3$  accounts for more than $90\%$, while the aggressive occupies less than $10\%$  for driver \#$9$ on highways. 

To further analyze the effects of personality (gender, age, driving experience) on the driving style mixtures, we split the driver participants into two gender groups (male and female), four age groups, and four driving experience groups. Experts scored the driving style of participants using the 5-Point Likert scale introduced in Section III-B. Fig.~\ref{fig: Driving topic distributions across different drivers} shows the distributions of the three specific driving styles with respect to personalities. The distribution differences are statistically significant ($p\le 0.01$) across the subgroups of each driver attribute (i.e., gender, age, driving experience, and driving style). 

\textbf{2) Females drive calmer than males.} Fig.~\ref{fig: Driving topic distributions across different drivers}(a) and (e) show that the percentage of aggressive style among females is much lower than that among males (while the proportion of the normal style is higher), indicating that females usually drive milder than males, which is consistent with the previous findings \cite{gwyther2012effect,sagberg2015review,laapotti2003comparison,zhao2012investigation,ericsson2000variability}. 

\textbf{3) Young drivers drive more aggressively.}  Fig.~\ref{fig: Driving topic distributions across different drivers}(b) and (f) show that the percentage of aggressive styles among drivers aged $21\sim 30$ is much higher than that among drivers aged $31\sim 50$ in both urban and highways, indicating that young drivers are more aggressive than others. However, drivers aged $51\sim 60$ drive calmly in urban and would become aggressive on highways. Some researchers also found that age appears to correlate negatively with aggressive driving by analyzing questionnaires of violations \cite{groeger1989assessing,dejoy1992examination,yagil1998gender}, and observed data about running red lights \cite{retting1996characteristics} or aggressive driving behaviors\cite{shinar2004aggressive}.

\textbf{4) Driving experience makes novice drivers more aggressive.} Previous studies shows that driving experience could affect aggressiveness \cite{sagberg2015review,lajunen2001aggressive,krahe2005predictors} by visual behavior \cite{mourant1970mapping,underwood2003visual} and operation skills \cite{lajunen1995driving,groeger1989assessing}. Fig.~\ref{fig: Driving topic distributions across different drivers} (c) and (h) reveal that drivers with less than five years of driving experience have a much higher proportion of calm style than other drivers, implying that unskilled drivers are more conservative. With the increase in the driving experience, the proportion of calm style decreases significantly, indicating that increasing driving experience can make drivers less conservative. 

\textbf{5) Drivers are recognizable in driving style distributions.} Fig.~\ref{fig: Driving topic distributions across different drivers} shows that the driving style distributions across driving styles are consistent with the driving styles labeled by the expert. For instance, aggressive drivers have the highest proportion of the aggressive topic among drivers, while calm drivers have the highest proportion of the calm topic among drivers. As for normal drivers, the proportion of the normal topic is highest among drivers.

\section{Conclusions}
This paper proposes a generic statistical framework to reveal three latent driving styles (i.e., calm, normal, and aggressive) that are shareable within and between individuals and verified using 100 drivers. We analytically build a learnable hierarchical latent model to capture the relationship between driving styles and driving behavior for individuals. The learned results indicate that a group of drivers shares the three driving styles, and the mixed weights of these driving styles lead to their diverse driving personalities.  For instance, aggressive drivers have a higher proportion than other drivers. 

The proposed hierarchical latent driving behavior model could facilitate the development of personalized intelligent driving systems by uncovering individuals' driving styles. A fixed sampling step was adopted for computational efficiency, but this might ignore the dynamic structure of driving behavior over time. Besides, determining the discrete word-like set of driving behavior in our proposed hierarchical latent model requires tedious testing. In future work, we plan to leverage factor analysis and temporal influences into the hierarchical latent model directly in a continuous space.


\appendices

\section{Likert Scale for Driving Style Questionnaire}
\label{app:likertscale}
We design a driving style questionnaire based on the Likert scale and finally classify the aggressiveness of driving styles into five levels, as shown in Table \ref{tab: Five aggressive levels of driver}.

\begin{table}[ht]
    \centering
    \caption{Five Aggressive Levels of Driving Behaviors}
    \label{tab: Five aggressive levels of driver}
    \begin{tabular}{p{2.5cm}p{5.5cm}}
    \hline\hline
     Levels  & Interpretation\\
     \hline
    Very poor aggressive & The driver strictly abides by safe driving regulations, averse to risk and stimulation.\\
    Poor aggressive      & The driver abides by safe driving regulations and tries to avoid risks. \\
    A bit aggressive     & The driver abides by safe driving regulations but occasionally pursues stimulation under the premise of safety.\\
    Aggressive           & The driver occasionally violates safe driving regulations, seeks stimulation, and can tolerate certain risks.\\
    Very aggressive      & The driver often violates safe driving regulations, pursues stimulation, and is willing to take higher risks.\\
    \hline\hline
    \end{tabular}
\end{table}

\section{Joint Distribution}
\label{sec:jointdistribution}

With Equation (\ref{eq: joint probability distribution}) and (\ref{Equation: joint distribution of behavior and style}), we can get
\begin{equation}
    \begin{aligned}
    p(\mathbf{z} \mid \boldsymbol{\alpha}) &=\int p(\mathbf{z} \mid \boldsymbol{\Theta}) p(\boldsymbol{\Theta} \mid \boldsymbol{\alpha})  \mathrm{d} \boldsymbol{\Theta} \\
    &=\int \prod_{d=1}^D p\left(\boldsymbol{\theta}_d \mid \boldsymbol{\alpha}\right) \prod_{n=1}^{N_d} p\left(z_{d, n} \mid \boldsymbol{\theta}_d \right)  \mathrm{d} \boldsymbol{\theta}_d \\
    &=\prod_{d=1}^D \frac{1}{\Delta(\boldsymbol{\alpha})} \int \prod_{k=1}^K \theta_{d, k}^{\alpha_k-1} \prod_{n=1}^{N_d} \theta_{d, z_{d, n}} \mathrm{d} \boldsymbol{\theta}_d \\
    &=\prod_{d=1}^D \frac{1}{\Delta(\boldsymbol{\alpha})} \int \prod_{k=1}^K \theta_{d, k}^{n_d^{z=k}+\alpha_k-1} \mathrm{d} \boldsymbol{\theta}_d \\
    &=\prod_{d=1}^D \frac{\Delta\left(\mathbf{n}_d+\boldsymbol{\alpha}\right)}{\Delta(\boldsymbol{\alpha})},
\end{aligned}
\end{equation}
where $\Delta(\cdot)$ denotes the Dirichlet operator, which will be canceled in the later process, $\mathbf{n}_{d} = \left\{n_d^{z=k}\right \}_{k=1}^{K}$ is the counts of each driving style shareable driving style $k$ for driver $d$.
Similarly, we get
\begin{equation} \label{Equation: joint distribution of behavior and style 2 -append}
    \begin{aligned}
    p(\mathbf{w} \mid \mathbf{z}, \boldsymbol{\beta}) &=\int p(\mathbf{w} \mid \boldsymbol{\Phi}, \mathbf{z}) p(\boldsymbol{\Phi} \mid \boldsymbol{\beta}) \mathrm{d} \boldsymbol{\Phi} \\
    &=\prod_{k=1}^K \frac{\Delta\left(\mathbf{n}_t+\boldsymbol{\beta}\right)}{\Delta(\boldsymbol{\beta})},
    \end{aligned}
\end{equation}
where $\mathbf{n}_{k} = \left\{ n_{z=k}^{i} \right \}_{i=1}^{M^m}$ is the counts of each driving word $w_i$ when driving style being $k$. Therefore, by Bayesian inference, we can get
\begin{equation} \label{Equation: joint distribution of behavior and style 3 -append}
    \begin{aligned}
     p(\mathbf{z}, \mathbf{w} \mid \boldsymbol{\alpha}, \boldsymbol{\beta})=& \iint p(\mathbf{w}, \mathbf{z}, \boldsymbol{\Theta}, \boldsymbol{\Phi} \mid \boldsymbol{\alpha}, \boldsymbol{\beta}) \mathrm{d} \boldsymbol{\Theta} \mathrm{d} \boldsymbol{\Phi} \\
     =& \int p(\mathbf{w} \mid \boldsymbol{\Phi}, \mathbf{z})  p(\boldsymbol{\Phi} \mid \boldsymbol{\beta})  \mathrm{d} \boldsymbol{\Phi} \\
     & \times \int p(\mathbf{z} \mid \boldsymbol{\Theta})  p(\boldsymbol{\Theta} \mid \boldsymbol{\alpha})  \mathrm{d} \boldsymbol{\Theta} \\
     &=p(\mathbf{w} \mid \mathbf{z}, \boldsymbol{\beta})  p(\mathbf{z} \mid \boldsymbol{\alpha}) \\
    &=\prod_{k=1}^K \frac{\Delta\left(\mathbf{n}_t+\boldsymbol{\beta}\right)}{\Delta(\boldsymbol{\beta})}  \prod_{d=1}^D \frac{\Delta\left(\mathbf{n}_d+\boldsymbol{\alpha}\right)}{\Delta(\boldsymbol{\alpha})}.
\end{aligned}
\end{equation}

\section{Conditional Distribution}
\label{sec: posterior distribution}

For the driving style $z_i$, which corresponds to $i$-th driving behavior fragments, the fully conditional probability has the following relationship via Bayesian formulation,
\begin{equation} \label{Equation: posterior probability of driving style i -append}
\begin{aligned}
    p\left(z_i \mid \mathbf{z}_{\neg i, \mathbf{w}}, \boldsymbol{\alpha}, \boldsymbol{\beta}\right)=& \frac{p(\mathbf{z}, \mathbf{w} \mid \boldsymbol{\alpha}, \boldsymbol{\beta})}{p\left(\mathbf{z}_{\neg i}, \mathbf{w} \mid \boldsymbol{\alpha}, \boldsymbol{\beta}\right)} \\
    =& \frac{p(\mathbf{z}, \mathbf{w} \mid \boldsymbol{\alpha}, \boldsymbol{\beta})}{p\left(\mathbf{z}_{\neg i},\left\{w_i, \mathbf{w}_{\neg i}\right\} \mid \boldsymbol{\alpha}, \boldsymbol{\beta}\right)} \\
    =& \frac{p(\mathbf{z}, \mathbf{W} \mid \boldsymbol{\alpha}, \boldsymbol{\beta})}{p\left(\mathbf{z}_{\neg i}, \mathbf{W}_{\neg i} \mid \boldsymbol{\alpha}, \boldsymbol{\beta}\right) \cdot p\left(w_i \mid \boldsymbol{\alpha}, \boldsymbol{\beta}\right)} \\
    \propto & \frac{p(\mathbf{z}, \mathbf{w} \mid \boldsymbol{\alpha}, \boldsymbol{\beta})}{p\left(\mathbf{z}_{\neg i}, \mathbf{w}_{\neg i} \mid \boldsymbol{\alpha}, \boldsymbol{\beta}\right)},
\end{aligned}
\end{equation}
where $\mathbf{z}_{\neg i} $ denotes all driving styles except for $z_i$, $\mathbf{w}$ are observed driving behavior fragments, and $\mathbf{w}_{\neg i} $ denotes the all other driving behavior fragments excluding the current one $w_i$. With (\ref{Equation: joint distribution of behavior and style 3 -append}) and (\ref{Equation: posterior probability of driving style i -append}), we get the fully conditional probability for $z_i =k$ as
\begin{equation} \label{Equation: posterior probability of driving style i 2 -append}
\begin{aligned}
     p\left(z_i=k \mid \mathbf{z}_{\neg i}, \mathbf{w}, \boldsymbol{\alpha}, \boldsymbol{\beta}\right) \propto 
         &\frac{n_k^i+\beta_i}{\sum_{j=1}^M\left(n_k^j+\beta_j\right)} 
        \\
         &\frac{n_d^k+\alpha_k}{\sum_{k=1}^K\left(n_d^k+\alpha_k\right)}. 
\end{aligned}
\end{equation}
Equation (\ref{Equation: posterior probability of driving style i 2 -append}) is used in Gibbs sampling to finally approximate estimate the posterior distribution of the driving styles among the test drivers.



\bibliographystyle{IEEEtran}
\bibliography{refs}

\begin{IEEEbiography}[{\includegraphics[width=1in,height=1.25in,clip,keepaspectratio]{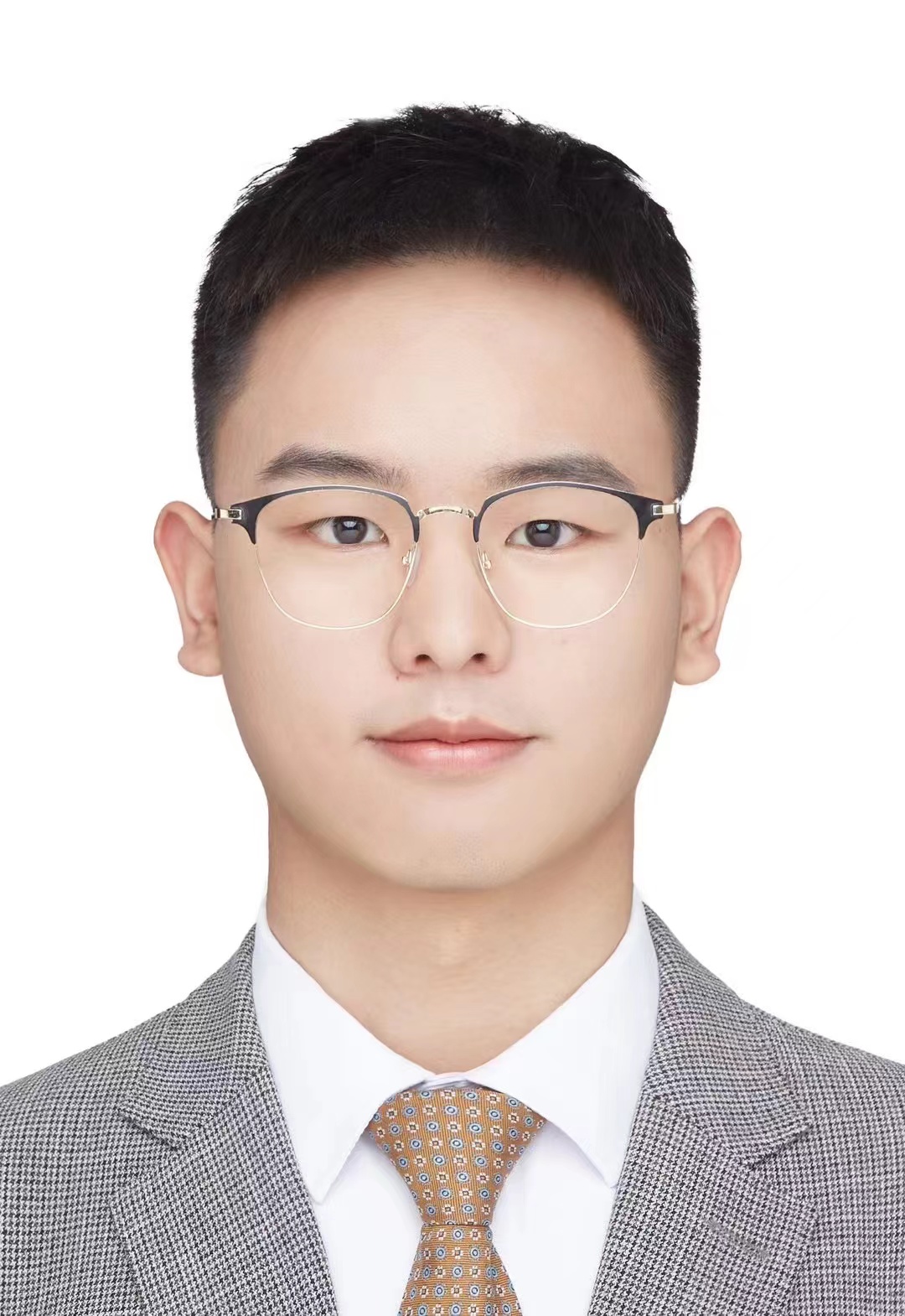}}]{Chaopeng Zhang} received B.S. degree in Mechanical Engineering from Beijing Institute of Technology, Beijing, China, in 2019, where he is currently working toward the Ph.D. degree in mechanical engineering. His research interests include human factors in intelligent vehicles, human driver model, driving style recognition, and driving intention recognition.
\end{IEEEbiography}

\begin{IEEEbiography}[{\includegraphics[width=1in,height=1.25in,clip,keepaspectratio]{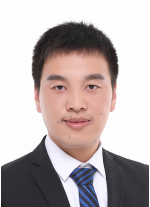}}]{WenshuoWang} (SM'15-M'18) received his Ph.D. degree in mechanical engineering from the Beijing Institute of Technology (BIT) in 2018.  Presently, he is a Full Professor at the School of Mechanical Engineering, BIT, Beijing, China. Prior to his role at BIT, he completed Postdoctoral fellowships at McGill University, Carnegie Mellon University (CMU), and UC Berkeley between 2018 and 2023. Furthermore, from 2015 to 2018, he served as a Research Assistant at UC Berkeley and the University of Michigan, Ann Arbor. His research interests focus on Bayesian nonparametric learning, human driver model, human–vehicle interaction, ADAS, and autonomous vehicles.
\end{IEEEbiography}

\begin{IEEEbiography}[{\includegraphics[width=1in,height=1.25in,clip,keepaspectratio]{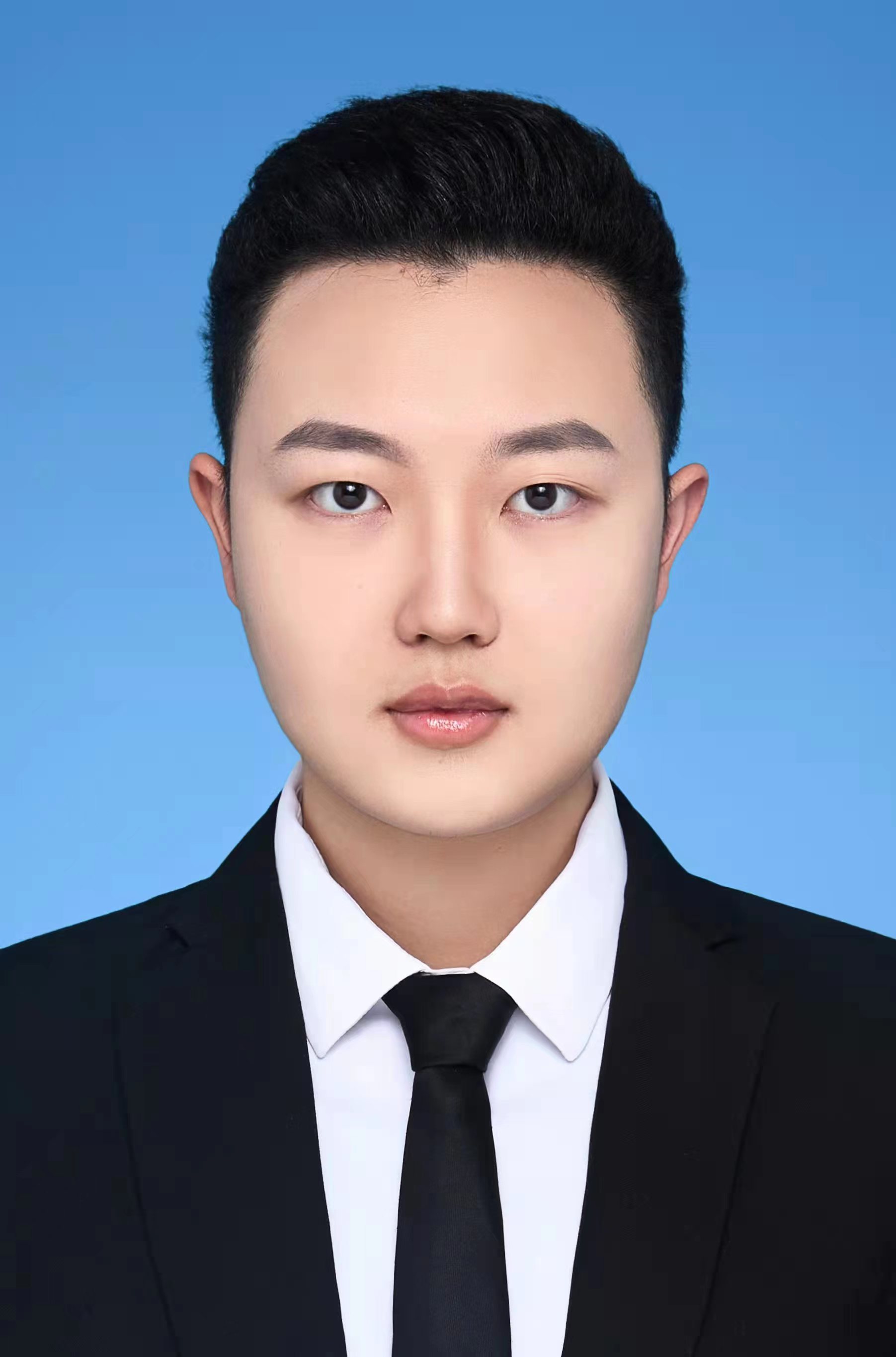}}]{Zhaokun Chen} received B.S. degree in intelligent automotive engineering from the Harbin Institute of Technology, Weihai, China, in 2022. He is currently working toward the M.S. degree in vehicle engineering with the Beijing Institute of Technology, Beijing, China. His research interests include driving behavior analysis, human driver model, driving intention recognition, and intelligent driving.
\end{IEEEbiography}

\begin{IEEEbiography}[{\includegraphics[width=1in,height=1.25in,clip,keepaspectratio]{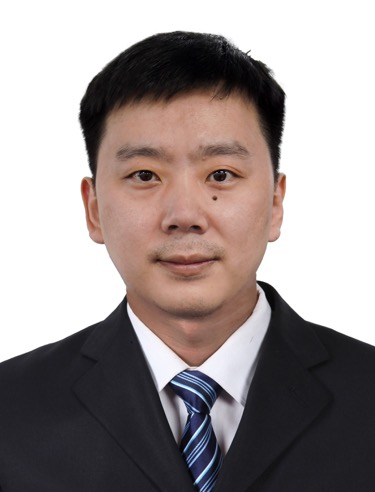}}]{JianZhang} Jian Zhang received the  M.S. degree in 2011 from the Jilin University, China. Currently, he is a senior director with Intelligent Connected Vehicle Development Institute at China FAW Group Co.Ltd., Jilin, China. His research interests include intelligent driving and vehicle control.
\end{IEEEbiography}

\begin{IEEEbiography}[{\includegraphics[width=1in,height=1.25in,clip,keepaspectratio]{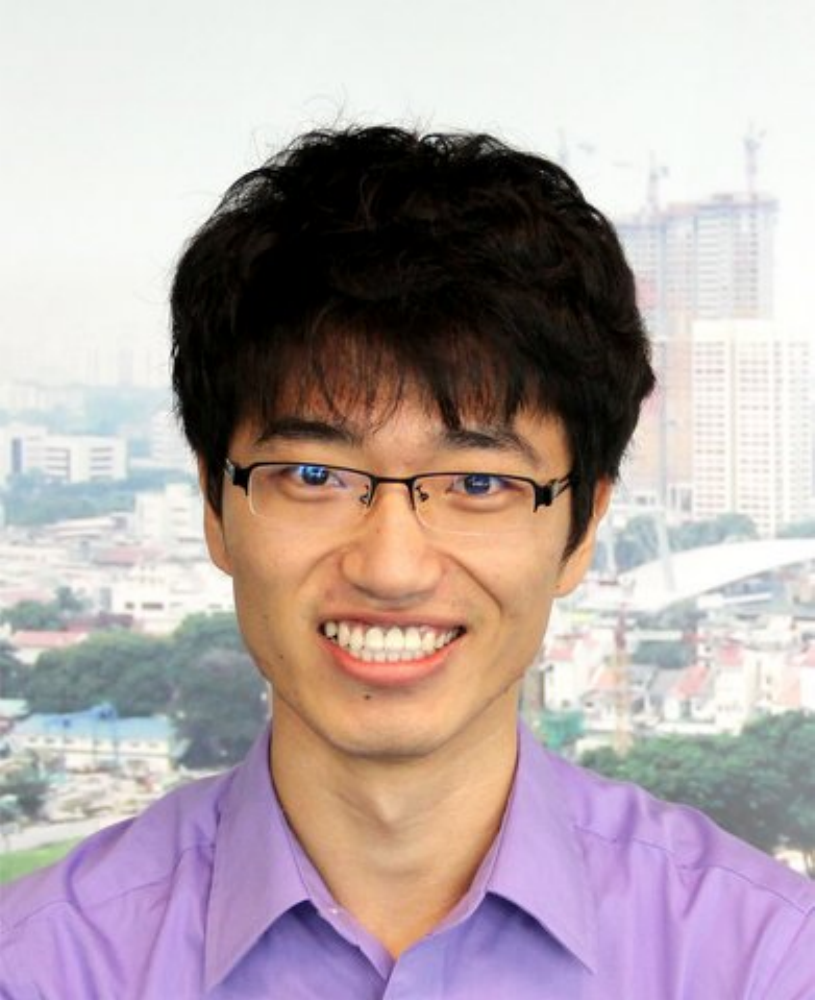}}]{Lijun Sun}
received B.S. degree in Civil Engineering from Tsinghua University, Beijing, China, in 2011 and Ph.D. degree in Civil Engineering (Transportation) from the National University of Singapore in 2015. He is currently an Assistant Professor with the Department of Civil Engineering at McGill University, Montreal, QC, Canada. His research centers on intelligent transportation systems, machine learning, spatiotemporal modeling, travel behavior, and agent-based simulation. He is an Associate Editor of Transportation Research Part C: Emerging Technologies. 
\end{IEEEbiography}

\begin{IEEEbiography}[{\includegraphics[width=1in,height=1.25in,clip,keepaspectratio]{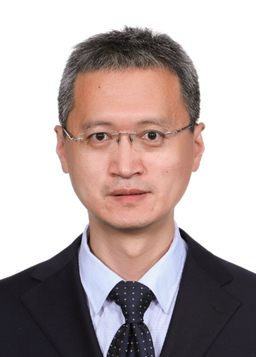}}]{JunqiangXi} received the B.S. degree in automotive engineering from the Harbin Institute of Technology, Harbin, China, in 1995, and the Ph.D. degree in vehicle engineering from the Beijing Institute of Technology (BIT), Beijing, China, in 2001. In 2001, he joined the State Key Laboratory of Vehicle Transmission, BIT. During 2012–2013, he made research as an Advanced Research Scholar in Vehicle Dynamic and Control Laboratory, Ohio State University, Columbus, OH, USA. He is currently a Professor and Director of Automotive Research Center in BIT. His research interests include vehicle dynamic and control, powertrain control, mechanics, intelligent transportation system, and intelligent vehicles.
\end{IEEEbiography}

\end{document}